\documentclass[pdflatex,sn-vancouver-num]{sn-jnl}

\usepackage{graphicx}%
\usepackage{multirow}%
\usepackage{amsmath,amssymb,amsfonts}%
\usepackage{amsthm}%
\usepackage{mathrsfs}%
\usepackage[title]{appendix}%
\usepackage{xcolor}%
\usepackage{textcomp}%
\usepackage{manyfoot}%
\usepackage{booktabs}%
\usepackage{algorithm}%
\usepackage{algorithmicx}%
\usepackage{algpseudocode}%
\usepackage{listings}%
\usepackage{array}
\theoremstyle{thmstyleone}%
\theoremstyle{thmstyletwo}%

\theoremstyle{thmstylethree}%

\begin{document}

\title[TR-RAG]{TA-RAG: Tone Awareness as a Design Imperative for Retrieval-Augmented Generation} 


\author*[1]{\fnm{Yong-Bin} \sur{Kang}}\email{ykang@swin.edu.au}

\author[1]{\fnm{Anthony} \sur{McCosker}}\email{amccosker@swin.edu.au}

\affil*[1]{Department of Media and Communication, Swinburne University of Technology, Melbourne, Victoria, Australia}


\abstract{
Retrieval-Augmented Generation (RAG) has become a robust architecture for grounding large language models (LLMs) in trusted knowledge. However, standard RAG systems exhibit a structural limitation: retrieved documents carry their own communication styles—professional jargon, formal tone, or academic writings—that shape the behavior of a RAG system before any tone instructions are processed, often causing the system to ignore user requests for a specific tone. We term this phenomenon \textit{contextual decoupling}, in which a system optimises for factual accuracy while remaining decoupled from the social or operational context of the recipient. Building on prior research in public health peer-support communities, we identify three communicative misalignment—linguistic, cognitive, and relational—that can persist even when retrieval is relevant and the generated response is factually accurate. We conceptualise these as failures of communicative transformation, which remain largely invisible to accuracy-centred RAG evaluation metrics. To address this gap, we propose \textbf{Tone-Aware RAG ({TA-RAG})}, a conceptual architectural framework that positions communicative alignment alongside factual accuracy as a core design objective.  TA-RAG operationalises four constraints—stigma-free language, readability alignment, recipient-sensitive adaptation, and empathetic framing—across the retrieval, context construction, generation, and constraint validation phases in the proposed RAG pipeline. We further highlight an evaluation agenda for jointly assessing factual fidelity and communicative alignment, and identify open challenges. We argue that tone awareness should be treated not as an optional refinement, but as a present design imperative for RAG systems operating in socially sensitive and high-stakes contexts.
}




\maketitle

\section{The Problem: Why Tone-awareness is Critical for RAG}\label{sec:intro}

RAG has established itself as a primary paradigm for deploying LLMs in knowledge-intensive domains. By coupling LLMs with curated knowledge sources, RAG offers a pragmatic solution to grounding responses in a trusted knowledge base \cite{gao2024rag}. In high-stakes fields, such as healthcare, legal practice, education, and public health domains—where factual reliability is paramount—this architecture represents a notable advancement. However, while RAG addresses the technical challenge of factual grounding, new forms of contextual decoupling are introduced in the failure to align with nuanced social or operational situations. We use the context-sensitive example of community healthcare and peer support to illustrate the need for better tone-aware control in RAG.

\subsection{Structural Conditioning at the Retrieval–Generation Interface}
What has not been solved through RAG is the structural flaw that the grounding mechanism itself introduces. Retrieved documents are not neutral information containers. They are textual artefacts that carry the distinct communicative norms of their origin contexts \cite{Yeh:2026}. Examples include the formal tone of a clinical protocol, the authoritative voice of a policy document, or the specialised language used in a medical field. 
When such documents are assembled into a prompt context, the generative model is strongly conditioned by their stylistic texture and token distributions before any runtime system instructions are parsed \cite{Yeh:2026, caut2026documenting}. Thus, in the standard RAG pipeline—comprising retrieval, context construction, and generation—the final communicative properties of a response are largely determined during retrieval, rather than generation.

As a result, soft tone instructions, using prompts, issued during generation—such as requests to ``be empathetic", ``use plain language", or ``avoid stigmatising terms"—must operate within a conditioning context already saturated by the linguistic distributions of the retrieved sources.
Recent research highlights that such generation-stage instructions are weakened when they compete with stronger stylistic or semantic signals in the retrieved context \cite{Liu23,ren-etal-2025-step}. This architectural boundary condition leads to what we term \textit{contextual decoupling}: a state in which the system accurately presents the retrieved fact but fails to align with the social or operational context in which that fact must be received. 
This limitation can be illustrated by sensitivities in communication within online mental health peer-support communities, where a technically accurate response may be communicated in an emotionally detached or overly formal way leading to audience-misalignment. In such situations, although the underlying information is correct, its delivery may fail to support, inform, or engage users effectively, and in some cases may undermine trust or cause harm \cite{Kang:2023,Kang:2022}. Because this problem arises from the composition of the context block itself, standard generation-stage prompting may not be sufficient to resolve it.

\begin{figure}[t!]
    \centering
\includegraphics[width=01\textwidth]{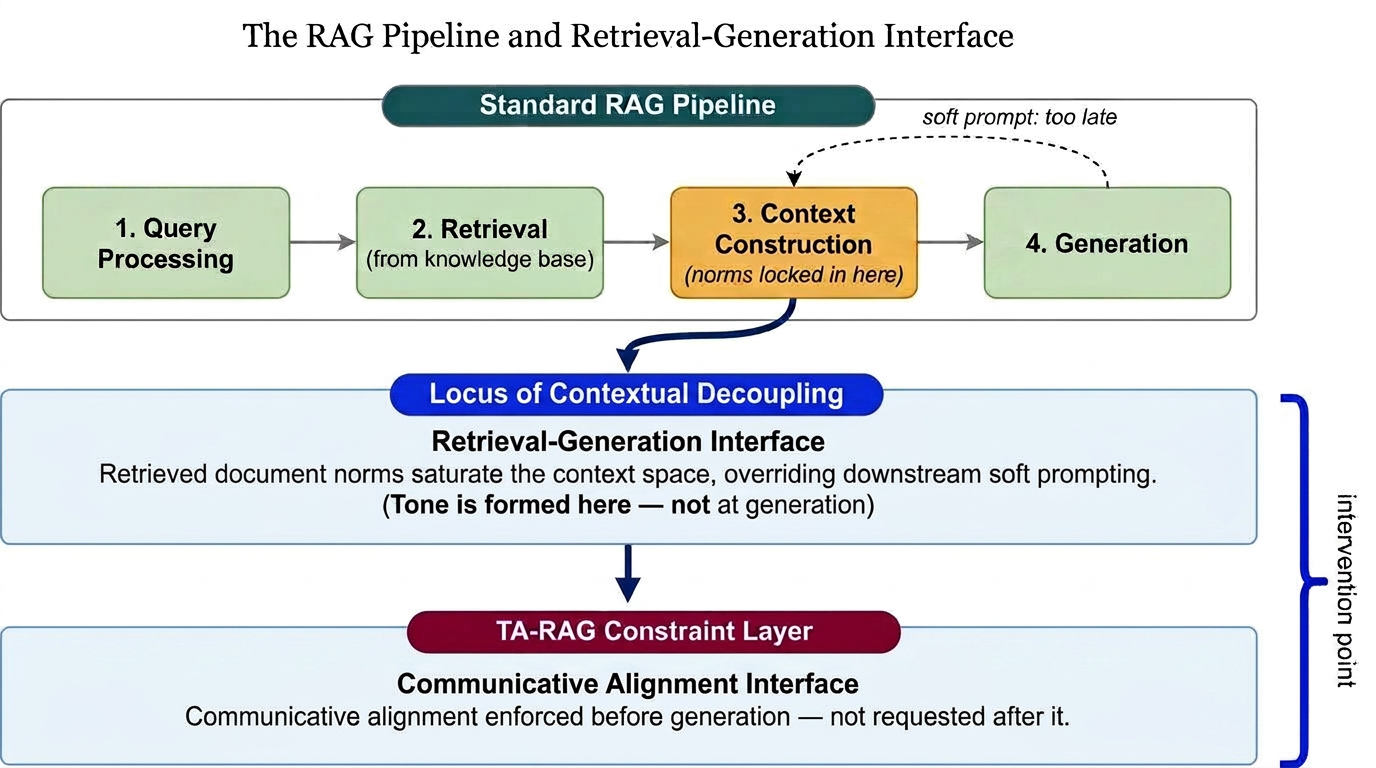}
\caption{
Contextual decoupling and the TA-RAG intervention. In Phase 3, retrieved document norms condition the context before generation. TA-RAG introduces a constraint layer at this interface to support communicative alignment.}    \label{fig:pre-ta-rag}
\end{figure}

Fig.~\ref{fig:pre-ta-rag} illustrates the contextual decoupling in the standard RAG pipeline. The upper tier traces the standard processing sequence. Stage (or Phase) 3 (context construction) represents the primary retrieval-generation interface because this is where retrieved documents are compiled into the prompt context and where their language patterns are \textit{locked in}. At this point, the style, formality, terminology, and tone of the retrieved sources begin to shape the context space before generation. The middle tier identifies this interface as the locus of the contextual decoupling. Although the system may retrieve factually relevant documents, it simultaneously imports the communicative styles embedded in the source texts. 
Because these features are already locked in at the interface, downstream soft prompting occurs too late, once the context space has been conditioned by the retrieved document norms. Later tone instructions have limited capacity to reliably reshape the final response.
To resolve this bottleneck, the lower tier positions the TA-RAG constraint layer as the necessary intervention point, enforcing communicative alignment at the interface before generation rather than attempting to adjust it after the style has been locked in. This can create greater control over the interaction, where organisations deploying the system can improve management of sensitive communication such as in healthcare contexts. 

\subsection{Three Communicative Failure Modes}

The structural weakness of RAG, particularly the locking-in of communicative norms at the retrieval--generation interface, produces three communicative failure modes. 
In each mode, the system may satisfy conventional criteria for a successful RAG response: the relevant document is retrieved, the content is factually accurate, and the user's query is addressed. Yet, the response can still fail on communicative grounds because it is misaligned with the audience, context, or relational expectations of the interaction. Table~\ref{tab:failure_modes} presents this typology:

\begin{table}[!t]
\small
\caption{Communicative failure modes in information-first RAG}
\label{tab:failure_modes}
\centering
\small
\begin{tabular}{>{\raggedright\arraybackslash}p{2.2cm}
                >{\raggedright\arraybackslash}p{3cm}
                >{\raggedright\arraybackslash}p{3cm}
                >{\raggedright\arraybackslash}p{3cm}
                }
\hline
\textbf{Failure Mode} & \textbf{Domain Anchor} & \textbf{Structural Cause} & \textbf{Evaluation Blind Spot} \\ \hline
\textbf{Linguistic Misalignment} & HIV peer-support: clinical sources reproduce ``HIV patient'' where person-first language is the norm & Retrieved institutional text imports deprecated terminology before generation instructions apply. & {Faithfulness satisfied:} Generated terms match the source text exactly. Linguistic or social harm is unmeasured. \\ \hline
\textbf{Cognitive Misalignment} & AI tutoring: expert-level textbook retrieved for a novice produces an explanation that alienates rather than scaffolds & Retrieved text’s reading level calibrated for source readership, not actual recipient & {Answer relevance satisfied:} Content is factually correct. Literacy and complexity mismatches are ignored. \\ \hline
\textbf{Relational Misalignment} & Mental health: clinical guideline retrieved for distressed user produces formally correct but emotionally detached response & Institutional communicative norms suppresses affective responsiveness before tone instructions apply& {Context precision satisfied:} The source document is relevant. Affective tone is not evaluated. \\
\hline
\end{tabular}
\end{table}

\textbf{Linguistic misalignment: stigma and terminological norm conflict.} 
Clinical and institutional documents carry terminological conventions appropriate to their source context but can be harmful in the target context. In HIV peer-support applications, retrieved clinical guidelines can contain ``HIV patient''—standard clinical usage, but explicitly discouraged in community settings (e.g., UNAIDS Terminology Guidelines \cite{unaids2024}) where ``person living with HIV" is the person-first standard. A RAG system  retrieves a factually correct document but reproduce terminology that is poorly suited to the user-facing context. Institutional terms may be acceptable in clinical or administrative records, yet feel distancing, judgemental, or stigmatising in peer-support contexts. The problem is therefore not the accuracy of the information, but the communicative effect of the terms used to present it. Such terminology can shape perceived judgement, willingness to engage with services, and psychosocial support experiences \citep{hem2024, Sharma2023HumanAI}. These choices should therefore be treated not merely as matters of style, but as clinically and socially consequential design properties.

\textbf{Cognitive misalignment: readability and expertise mismatch.} 
Professional texts encode reading levels calibrated to their source readership—a clinical protocol for physicians, a textbook for advanced students, a policy document for administrators. When retrieved for a non-expert user, a patient, or a novice learner, this level is  mismatched to the actual recipient. The content is correct and relevant, but the RAG pipeline has no mechanism to model or close this communicative distance. Such mismatches substantially reduce comprehension, engagement, and behavioural outcomes \cite{Bol:2020}. In AI tutoring, LLM-based systems have been shown to prioritise direct answers over scaffolded, dialogic interaction \cite{beale2025dialogic}—a tendency compounded when retrieved materials encode expert-level assumptions about the reader.

\textbf{Relational misalignment: tone conflict and the empathy gap.}
Institutional sources sound authoritative and neutral. In peer-support and mental health contexts, users seek acknowledgment of their emotional state alongside—often before—informational content. A RAG system retrieving clinical guidelines for a distressed user inherits the source's formal, objective, emotionally flat tone, suppressing the warmth the interaction requires before the model generates a word. This has been identified as the ``empathy gap" in conversational AI \cite{Liu2025}. Relational qualities such as warmth and responsiveness are functional determinants of sustained human-AI engagement, not stylistic features \cite{Bickmore2005}, and empathic communication in peer-support contexts produces measurable improvements in user trust, engagement, and outcomes \cite{Sharma:2021, Sharma2023HumanAI}. The RAG pipeline offers little support for generating these qualities when retrieved sources work against them.

\subsection{Why current RAG metrics cannot capture these failures}
These failure modes highlight an important property: they are difficult to detect using standard RAG evaluation metrics. This is structural rather than incidental, and it helps explain why tone-related issues in RAG outputs have persisted even as awareness of tone-related concerns in LLM outputs has grown.
Standard RAG evaluation frameworks—RAGAS \cite{es-etal-2024-ragas} and comparable approaches—primarily assess faithfulness (\textit{does the response accurately reflect retrieved content?}), answer relevance (\textit{does it address the query?}), and context precision (\textit{are the retrieved documents appropriate to the query?}). These are valuable factual assessments. However, in each of the failure modes, a response can score well on all three while still exhibiting the communicative issue. A response reproducing ``HIV patient" from the retrieved source remains faithful to it. An expert-level explanation drawn from retrieved textbook content remains relevant. Vernacular, culturally distinct and context sensitive modes of communication are easily lost in the interaction. A formally correct response to a distressed user, based on a precise clinical guideline, remains informatively appropriate—the content is suitable—but the tonal dimension falls largely outside of the metrics.

This gap reflects an underlying orientation in how RAG quality has typically been conceptualised: factual accuracy has served as the primary, criterion against which RAG outputs are evaluated, and evaluation frameworks have developed accordingly. When communicative alignment is not measured, it becomes harder to monitor or improve systematically. When it is not reported, it is less likely to be prioritised. 

Addressing this gap requires two complementary developments. Section 2 introduces TA-RAG, a framework that articulates what communicative alignment means in RAG systems and how it can be incorporated as a design consideration alongside factual accuracy. Because putting such a framework into practice depends on the ability to assess whether communicative goals are being met, Section 3 proposes an evaluation agenda that we view as integral to the TA-RAG proposition. Section 4 then considers the broader implications of this perspective and outlines directions for future research.

\section{TA-RAG: Framing, Architecture, and Operationalisation}\label{sec:ta-definition}

The structural diagnosis detailed in Section \ref{sec:intro} suggests that attempting to govern the output tone exclusively through prompts at the generation phase represents an incomplete solution. Because the communicative conditioning introduced by retrieved documents can structurally precede and  override generation-level instructions, interventions are more effectively implemented upstream within the retrieval and context construction phases. We therefore propose reframing tone-awareness as an essential architectural design criterion equal in standing to factual validity, enforced across all phases of the pipeline. 

Conventional RAG frameworks optimise primarily for a single quality dimension—namely, how faithfully and relevantly the output reflects retrieved factual content. Distinctively, TA-RAG  takes a dual-optimisation approach. Outputs within TA-RAG  must simultaneously satisfy factual precision and target communicative parameters. Consequently, a factually accurate but communicatively misaligned response is not an admissible output—it must be revised or, in high-stakes contexts, escalated for human review.

This approach aligns with the concept of interdiscursivity \cite{Fairclough2010}, which explains how text distributions draw on and transform the conventions of alternative discourse genres. For example, if a RAG pipeline was set up to retrieve an authoritative medical protocol to answer a query for a healthcare-oriented chatbot it intersects two distinct communicative contexts: the formal genre of clinical documentation and the informal genre of community-based support for people not medically trained. Retrieved documents are fundamentally communicative artifacts whose intrinsic properties—such as institutional distance or highly specialised style—persist into and condition the final generated text. Stylometric analyses confirm that these linguistic parameters independently modulate reader engagement and trust, irrespective of underlying semantic accuracy \cite{Kang:2022, Kang:2023}.  TA-RAG  provides the architectural mechanisms to systematically govern this interdiscursive transformation.

\subsection{The Four Communicative Constraints}

TA-RAG  operationalises communicative alignment through four distinct system-level constraints, each directly mapping to a corresponding failure mode identified in Section 1 and measuring the tonal dimensions of outputs:

\begin{itemize}
    \item \textbf{Stigma-free language} addresses linguistic misalignment. It requires generated responses to follow domain-specific terminology standards, for example UNAIDS Terminology Guidelines \cite{unaids2024} in HIV contexts, as well as equivalent frameworks for mental health, disability, and neurodiversity. This constraints can be satisfied relatively easily because authoritative terminology guidelines are available and compliance can often be assessed through rule-based or deterministic methods. Its significance extends beyond technical correctness. Terminological choices can influence perceived dignity, trust, and willingness to engage with services, with implications emphasised in health and community settings \cite{hem2024, hiv2023stigma}.

    \item \textbf{Readability alignment} addresses cognitive misalignment by calibrating the lexical and structural complexity of the output to the recipient profile \cite{Tran2025MedReadCtrl}, rather than allowing the response to inherit the reading level of the retrieved source. Traditional readability metrics, such as Flesch-Kincaid and SMOG, can be used complementarily with recent zero-shot LLM-based Automatic Readability Assessment (ARA) \cite{grossman:2026}, which estimates contextual readability by assigning difficulty scores to generated text. These indicators could offer measurable thresholds against which outputs can be assessed. The key architectural challenge is not only to measure readability, but also to intervene when the retrieved source text is too complex: the TA-RAG pipeline must actively counteract the complex textual properties imported by professional source documents and substitute structural scaffolding appropriate to the recipient's literacy level.
    
    \item \textbf{Recipient-sensitive adaptation} addresses the audience-facing dimension of relational misalignment. It requires responses to be shaped around the communicative needs of the actual recipient, rather than the generic reader implied by the retrieved source. This involves modelling relevant recipient characteristics (e.g., role, domain expertise, cultural context and communicative objective) and using such information to guide language style, technical depth, and explanatory scaffolding. For example, the same retrieved mental health guideline may need to be explained differently to a clinician, a peer-support worker, and a distressed service user: the factual basis remains the same, but the level of detail, tone, and supportive framing should differ. Research on tailored communication suggests that aligning message characteristics with recipient attributes can improve comprehension, engagement, and behavioural outcomes \cite{Mia:2017, Bol:2020, Lapinski:2025}.

    \item \textbf{Empathetic framing} addresses the affective dimension of relational misalignment. It requires that responses appropriately respond to the emotional context of the interaction, not merely its informational content. A precise distinction is necessary: TA-RAG does not attempt to make AI systems feel empathetic—which is both technically infeasible and ethically contested. It requires that AI systems communicate appropriately in emotionally sensitive contexts: recognising affective cues in user input and adjusting communicative stance accordingly, for instance, by acknowledging distress before delivering guidance. 
    Prior research shows that this kind of affective alignment is trainable, measurable, and consequential for user outcomes \cite{shen2024empathy, Sharma2023HumanAI}. The work on the illusion of empathy confirms that users reliably detect its absence, and that detection has measurable consequences for trust and re-engagement \cite{Liu2025}.
    
\end{itemize}

\subsection{TA-RAG Architecture and Operationalisation}

\begin{figure}[t!]
    \centering
\includegraphics[width=01\textwidth]{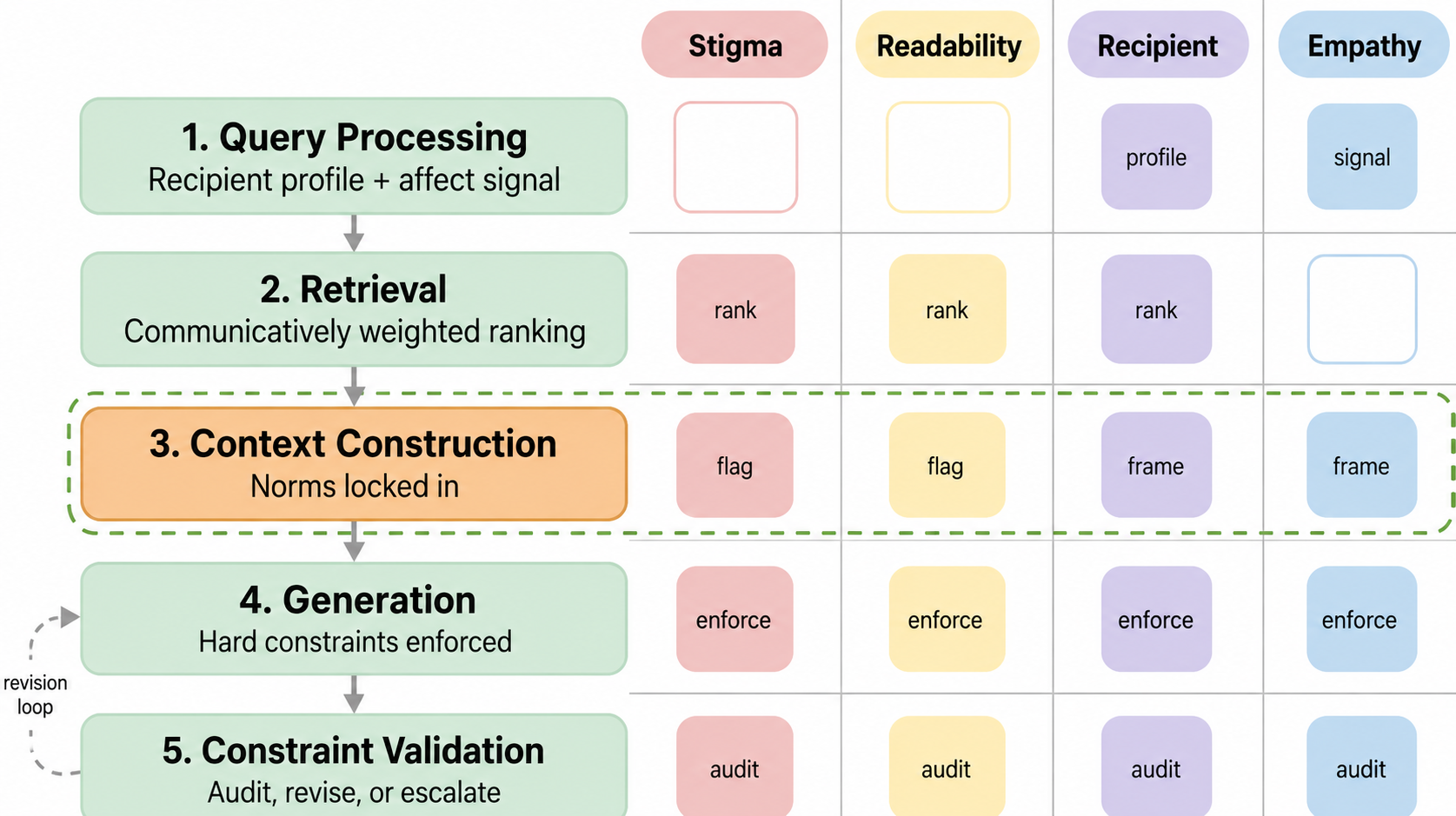}
\caption{TA-RAG constraint placement across the RAG pipeline. Each cell specifies the operational verb (profile, signal, rank, flag, frame, enforce, audit) associated with a constraint at that pipeline phase. Empty cells indicate at which a constraint is not yet operative.}
\label{fig:tarag_arch}
\end{figure}

The TA-RAG architecture is presented in Fig.~\ref{fig:tarag_arch}, which maps
the four communicative constraints onto the RAG pipeline, providing the architectural constraint layer posed in Fig.~\ref{fig:pre-ta-rag}. 
The architecture is structured as a 5-phase vertical pipeline. 
On the left, the core processing phases are shown; on the right, a matrix of cells indicate the operational status of each constraint  at each phase. Each cell specifies the action a constraint performs at that phase, while an empty cell indicates that the constraint is not yet operative.

The architecture introduces three structural additions to the standard RAG pipeline. First, query processing (Phase 1) is expanded into differentiated input extraction, separating recipient information into a \textit{profile} and an affective \textit{signal}. Second, communicative constraints are injected as early in the pipeline as possible (Phases 2--4), where they can proactively guide system behaviour, rather than being deferred entirely to generation. Third, a constraint validation phase (Phase 5) audits the generated response and triggers targeted revision where a constraint is not met, avoiding full regeneration where focused repair is sufficient.

The guideline resources used by TA-RAG are not generated ad hoc by the model during interaction. They are specified in advance during system design and may include domain-specific terminology guidelines, stigma-free language resources, readability thresholds, recipient-adaptation rules, and empathy or peer-support communication rubrics. Which resources are activated depends on the recipient profile and affective signal extracted during query processing. For example, a patient or peer-support user may activate person-first terminology, plain-language, and empathetic framing guidelines, whereas a professional user may permit more technical terminology and greater informational density. These resources should be defined by system designers in consultation with domain experts, ethics or governance bodies, and, where appropriate, representatives of the affected user community.

The operational flow can be described as a sequence of phase-specific interventions:

\begin{enumerate}
    \item \textbf{Query processing.} A communicative parsing phase produces two outputs: a recipient \textit{profile}, capturing user attributes such as role, expertise, or likely information needs, and an affective \textit{signal}, representing distress, vulnerability, or emotional urgency expressed in the query. These outputs propagate through subsequent phases and guide how constraints are applied.
     
    \item \textbf{Retrieval.} Semantic similarity ranking is complemented by three communicative ranking signals. A terminology signal gives lower priority to sources that use deprecated or non-preferred terms; a readability signal favours sources whose complexity is closer to the inferred or specified target reading level; and a role-relevance signal weights documents according to the recipient profile. The empathetic framing constraint is not operative at this phase, since affective appropriateness primarily concerns how retrieved content is expressed rather than which documents are selected.
    
    \item \textbf{Context Construction.} This phase forms the retrieval--generation interface and is the first point at which all four constraints become active together.  While Phase 4 enforces constraints during generation and Phase 5 audits the output afterward, Phase 3 fundamentally shapes the communicative character of the response. Source material remains individually addressable here, allowing specific passages to be annotated before they are combined into the conditioning context for generation.  The stigma-free language and readability constraints are \textit{flagged}: passages containing non-preferred terms are annotated with suggested substitutions, and passages exceeding the target reading level are annotated with simplification guidance. At the same time, the recipient adaptation and empathetic framing constraints \textit{frame} the context by attaching recipient-related metadata and, where relevant, instructions for acknowledging the user's emotional state. This convergence of all four constraints makes context construction the architectural focal point of TA-RAG.
   
    \item \textbf{Generation.} The model receives the prepared context together with constraint-related instructions assembled during \textit{Context Construction}. These instructions draw on the recipient profile, affective signal, flagged passages, and relevant guideline resources. The model is then expected to \textit{enforce}: avoiding flagged terms in favour of preferred alternatives, aiming for the specified reading-level target, adjusting explanatory depth according to the recipient profile, and including an acknowledgement of the user's emotional state where the affective signal indicates this is appropriate.
   
    \item \textbf{Constraint Validation.} The generated output is then \textit{audited} against all four constraints. Terminology checks identify any remaining non-preferred terms and trigger targeted revision where needed. Readability is assessed using standard metrics, with iterative simplification applied if the target level is not met, capped at three cycles. Recipient adaptation and empathetic framing are assessed through rubric-based scoring. Cases that remain unresolved after automated revision can be directed to human-in-the-loop review.
\end{enumerate}

\subsection{What distinguishes TA-RAG from adjacent approaches}

Adjacent research typically addresses personalisation, readability, stigma-free language, or empathetic framing as isolated, localised controls. TA-RAG distinguishes itself by treating these dimensions as coordinated constraints distributed across the full 5-phase pipeline, as elaborated below.

\begin{enumerate}
    \item  \textbf{From persona-conditioned personalisation:} Existing personalisation methods rely on retrieving user profiles (i.e., user historical data) to alter language, framing, or emphasis at the prompt level \citep{salemi:2024, chong:2025}. This can be problematic when users switch between personal and professional contexts of use, for instance. TA-RAG elevates recipient adaptation from a prompt-level condition to a pipeline-level concern: recipient signals guide retrieval, context construction, generation, and validation, rather than functioning only as prompt- or profile-level conditioning.

    \item  \textbf{From readability-control models:} Readability-aware prompting steers output complexity towards target reading levels \citep{Tran2025MedReadCtrl}, but these approaches frequently miss targets, and standalone simplification can degrade factual accuracy \citep{qiu:2025}.  TA-RAG therefore treats readability not only as a generation-phase objective but as a pipeline constraint: retrieved sources can be ranked by proximity to the recipient's target level, context can be annotated with readability requirements, and final outputs can be validated against both readability and factual-retention criteria.

    \item  \textbf{From prompt-based stigma mitigation:} Recent work on stigmatising clinical documentation shows that subtle linguistic framing can influence LLM clinical decision-making and simulated attitudes toward patients, while prompt-based mitigation offers only partial protection \citep{Huang:2026}. TA-RAG positions stigma-free language as a phased control rather than an output filter: retrieval can downweight sources containing deprecated terminology, context construction can annotate preferred substitutions, generation can encode prohibited terms alongside alternatives, and validation can trigger targeted revision where non-preferred language persists.

    \item  \textbf{From isolated empathic rewriting:} Research on empathic rewriting and Human-AI support systems demonstrates that empathy can be computationally modelled and improved through rewriting, while requiring careful attention to specificity, fluency, safety, and human agency \citep{Sharma:2021, Sharma2023HumanAI}. The limitation is that such work operates at the output level, without addressing how the register and tone of retrieved source material may suppress affective responsiveness before generation begins. TA-RAG operationalises empathetic framing as a pipeline-wide constraint: affective cues are detected during query processing, framing instructions are carried through context construction and generation, and the response is audited for appropriate acknowledgement and tonal alignment.
\end{enumerate}

The primary distinction of TA-RAG's is its shift from isolated stylistic management to distributed integration, placement, and verification. It offers a systematic form of institutional or organisational control and user agency. TA-RAG distributes constraint handling across the phases where each property is most actionable. This operationalises the claim that the communicative properties of a RAG response accumulate across the pipeline. Its validation phase further shifts communicative alignment from a requested stylistic preference toward an auditable pipeline property: outputs are checked against each constraint, revised when thresholds are not met, and escalated to human review where appropriate. This distinction is particularly important in domains where misalignment can affect safety, trust, and ongoing engagement.
\section{Evaluation agenda and criteria for Communicative Alignment in TA-RAG}\label{sec:ta-eval}

The TA-RAG proposition holds that communicative alignment should be enforced as an architectural constraint. Enforcement, however, depends on verifiability, which in turn depends on measurement. The evaluation agenda is therefore not an optional extension of the framework, but a necessary part of it: without reliable metrics, the constraint validation phase has limited means to determine whether a response satisfies a given constraint, making system quality difficult to ensure in practice.

At present, such a measurement framework does not exist in an integrated form. Standard RAG evaluation practice has largely prioritised faithfulness, answer relevance, and context precision, while empathy, readability, and terminological appropriateness remain comparatively underexamined as reportable quality dimensions. Advancing evaluation standards is therefore as important as advancing architectural frameworks. We highlight this agenda below by considering each constraint in turn, before addressing how factual accuracy and communicative alignment might be assessed jointly.

\subsection{Constraint-specific criteria}
Stigma-free language may be the most readily evaluable of the four constraints, since authoritative reference glossaries already exist and compliance can be assessed in a largely deterministic manner. Resources such as  UNAIDS Terminology Guidelines \citep{unaids2024} and mental health and disability style guides, provide reference vocabularies against which outputs can be compared. A meaningful open challenge concerns context-sensitive disambiguation: the term ``patient,'' for instance, is generally appropriate in clinical framing but less suitable in peer-support contexts, suggesting that \textit{classifiers} will need to assess contextual appropriateness rather than relying on term presence alone. Human expert review remains valuable for ambiguous cases, and developing annotated datasets to support context-sensitive stigma classifiers represents a near-term priority for the community.

Readability alignment benefits from established, deterministic metrics—Flesch-Kincaid, SMOG, and related variants—that assess complexity through linguistic features such as syllable count and sentence length. Recent LLM-based Automatic Readability Assessment \citep{grossman:2026} can offer a complementary approach, estimating contextual readability from a model's probability distribution over difficulty scores rather than from surface form alone—a useful corrective in cases where structurally simple text remains conceptually dense, or vice versa. Because appropriate simplification thresholds are likely to vary by domain, their calibration would benefit from input by domain experts rather than being set uniformly—itself an open question worth further attention.

Recipient adaptation can be assessed through expert rubric evaluation of language style, technical depth, and explanatory scaffolding. A complementary approach would draw on domain-specific practice guidelines, such as  NAPWHA Australian Peer Support Standards \citep{napwha2020peerstandards}, to derive structured and checkable criteria. These criteria could cover aspects such as preferred modes of address, the sequencing of technical information, and the extent to which a response acknowledges the recipient’s situation or needs. Applied alongside expert judgement, such rules could improve consistency without replacing rubric-based assessment entirely.
Deriving these programmatic rules from unstructured, text-based documents remains an open research direction. Because clinical and peer standards are typically written in discursive language for human practitioners rather than as formal specifications for automated systems, extracting verifiable criteria while preserving their underlying clinical nuance introduces non-trivial methodological challenges.  Text-heavy practitioner guidelines are common across high-stakes domains, including mental health and disability support, suggesting that this translation challenge extends beyond any single document or application context.

Empathetic framing is among the most difficult constraints to evaluate because it depends strongly on context. The aim is not to assess whether an AI system is genuinely empathetic, but whether its response uses an affective language that is appropriate for the situation. This distinction is important: claims about affective authenticity raise ethical concerns for AI systems, whereas affective appropriateness can be evaluated through observable features of the response. Evidence that users can distinguish genuine from formulaic affective engagement, with implications for trust \citep{Liu2025}, highlights the practical importance of this issue. Related work on reinforcement-learning approaches to empathic rewriting and feedback \citep{Sharma2023HumanAI, Sharma:2021} further suggests that judgments about empathic quality can be elicited with reasonable reliability. The remaining challenge is to distinguish responses that acknowledge the user's situation in a contextually appropriate way from responses that merely use surface-level empathetic phrases without fitting the situation.

\subsection{Joint evaluation: the accuracy–alignment frontier}

Constraint-specific metrics are valuable, but unlikely to be sufficient on their own. A complete evaluation framework should assess factual accuracy and communicative alignment jointly on the same query--response pairs, since improving one communicative dimension may place pressure on another or on factual fidelity. This concern applies especially when the ``constrain validation phase" triggers revision. Whether the system substitutes terminology, simplifies a passage, reframes content for a different recipient, or adds an affective acknowledgement, the revised output should preserve the substantive content of the original response. A revision that satisfies one constraint while distorting or omitting important information would represent a regression, even if it scores well on the targeted constraint.

Semantic preservation metrics, including embedding-based similarity measures and BERTScore \citep{zhang2020bertscore}, can offer a practical way to guard against this risk across all four constraints. Embedding-based measures estimate whether two texts remain close in meaning, while BERTScore compares contextual token-level similarity between a revised output and a reference text. By comparing a response before and after revision, the validation phase can assess whether the underlying message has been preserved while the communicative property is improved. In this arrangement, constraint-specific metrics verify that the relevant communicative requirement has been met, while semantic preservation metrics check that meeting it has not changed the response's substantive meaning.

Building this evaluation framework would require benchmark datasets that pair factual accuracy and communicative alignment annotations across domains such as healthcare, education, and peer support; reporting protocols that treat both dimensions as standard evaluation criteria; and test sets designed to surface tensions between retrieved source material and recipient context. This is a community-level undertaking comparable in ambition to retrieval benchmarks such as BEIR \cite{thakur2021beir}. TA-RAG's four-constraint structure, combined with a constraint-agnostic semantic preservation check, can provide a strong basis for benchmark design. However, constructing and validating the required datasets, metrics, and reporting standards remains an important open direction for treating communicative alignment as a measurable dimension of RAG system quality.

\section{Implications and Research Roadmap}

\subsection{Governance of the accuracy–alignment trade-off}

Communicative alignment and factual fidelity are not always easily compatible. A credible account of TA-RAG should therefore address this tension directly. Simplifying a clinical explanation to meet a target reading level may reduce important nuance; softening risk-related information to reduce distress may, in some contexts, weaken its practical urgency. These are familiar tensions in health communication and medical ethics, not concerns unique to TA-RAG.

TA-RAG does not aim to resolve such trade-offs universally. Its contribution is to make them visible, measurable, and open to deliberate calibration, rather than leaving them as implicit outcomes of default system behaviour. Domain-specific thresholds at the ``constraint validation phase" can provide one possible mechanism: near-zero tolerance for stigmatising terminology in peer-support contexts, greater flexibility for simplification in patient education, or reduced emphasis on empathetic framing in time-critical emergency triage, where informational urgency may reasonably take priority. Treating these thresholds as explicit and adjustable allows trade-offs to be governed with input from relevant domain expertise. Input could include, for instance, tone rich knowledge sets including lived experience, forms of organisational tacit knowledge, or vernacular ways of communicating. 

An important asymmetry may also arise in practice. In peer-support contexts, a response that is slightly less precise but substantially more appropriate in tone and terminology may serve the user better than one that is technically precise but stigmatising or affectively misaligned. Information-first designs often default toward the latter, partly because communicative quality is less often measured and more difficult to operationalise. Making this trade-off explicit is therefore a necessary step toward deciding how it should be managed in different settings.

\subsection{Open research challenges}

Realising TA-RAG more fully will require progress in several areas. We present these as research challenges for the broader community:

\begin{enumerate}
    \item First, retrieval mechanisms need to become more sensitive to communicative style as well as semantic relevance. Current dense retrieval systems are mainly optimised for topical similarity. Extending them to account for language, preferred terminology, and reading-level compatibility may require style-aware embedding spaces, communicative metadata at indexing time, or new training resources.

    \item Second, more robust automated classifiers are needed for constraints that still depend heavily on expert judgement to address varied lived experience and improve recipient adaptation and empathetic framing. This requires annotated datasets and, for recipient adaptation, methods for translating practice guidelines into structured evaluation criteria.

    \item Third, the field needs joint benchmarks that assess factual accuracy and communicative alignment on the same datasets. Such benchmarks would make it possible to study when alignment interventions preserve, strengthen, or place pressure on factual fidelity. This is a substantial community-level task, comparable in ambition to established retrieval evaluation efforts.

    \item Fourth, governance remains an open challenge. Thresholds for ``constraint validation phase" must be set, justified, reviewed, and revised as evidence accumulates in collaboration with domain experts and people with lived experience and cultural or tacit knowledge. These decisions require collaboration among NLP researchers, domain practitioners, and the communities most affected by these systems.
\end{enumerate}

\subsection{Towards communicatively aligned RAG systems}

This perspective paper has argued that communicative alignment should be treated as a core design consideration in RAG systems, alongside factual accuracy. This is not a claim that current RAG architectures are without value, nor that prompting and post-processing approaches are without merit—they remain useful tools. It is, rather, a suggestion that these tools may be insufficient on their own in domains where the communicative dimension of a response materially affects how it is received, and that a more architecturally integrated approach, one that enables scrutiny of tonal sensitivities and solutions to contextual decoupling, offers a complementary path forward.

TA-RAG represents an initial attempt to specify what such integration might look like in practice: where in a RAG pipeline communicative constraints might be usefully introduced, how they might be verified, and how the resulting trade-offs might be governed transparently. Considerable work remains in developing the evaluation framework, retrieval mechanisms, and governance processes that this approach requires. We see this as a crucial research programme and one with particular relevance to domains—healthcare, mental health support, and education, among them—where the manner in which information is communicated can matter as much as its accuracy.

\section*{Declarations}

\subsection*{Funding}
This work is supported by the ARC Centre of Excellence for Automated Decision-Making and Society (ADM+S), funded by the Australian Government through the Australian Research Council (CE200100005).

\subsection*{Author contributions}
Y.K. and A.M. conceptualised the paper and developed the manuscript. Y.K. led the overall framing of the paper, literature analysis, development of the conceptual architecture, and drafting of the manuscript. A.M. contributed to the theoretical framing, interpretation, and revision of the manuscript. Both authors reviewed and approved the final manuscript.

\subsection*{Ethics approval}
Ethics declaration: not applicable. 

\subsection*{Consent to participate}
Consent to participate declaration: not applicable.

\subsection*{Consent to publish}
Consent to publish declaration: not applicable.

\subsection*{Clinical trial registration}
Clinical trial number: not applicable.

\bibliography{ta-rag}

\end{document}